\documentclass{article}

\usepackage[sglblindworkshop, final]{neurips_2026}
\workshoptitle{AI Foundations for Power Grids}

\usepackage[utf8]{inputenc}
\usepackage[T1]{fontenc}      
\usepackage{graphicx}      
\usepackage{amsmath}               
\allowdisplaybreaks
\usepackage{amssymb}               
\usepackage{amsfonts}              
\usepackage{amsbsy}
\usepackage{bbm}
\usepackage{tikz}                  
\usepackage{multicol}
\usepackage{graphicx}
\usepackage{threeparttable}
\usepackage{soul,derivative,stmaryrd,mathtools,enumerate,enumitem,mathtools}
\usepackage{comment}
\usepackage{algorithm}
\usepackage{algorithmicx}
\usepackage{algpseudocode}
\usepackage[colorlinks=true,citecolor=blue]{hyperref}
\usepackage{xfrac}
\usepackage{wrapfig}

\usetikzlibrary{arrows,positioning,shapes}
\usetikzlibrary{shapes.multipart}
\usetikzlibrary{arrows.meta}
\usetikzlibrary{decorations.pathreplacing}
\usetikzlibrary{arrows,intersections}
\usetikzlibrary{shapes,arrows}
\usetikzlibrary{fit,positioning}

\newcommand{\mbb}[1]{\mathbb{#1}}

\newcommand{\mrm}[1]{\mathrm{#1}}

\newcommand{\R}{\mbb{R}}

\newcommand{\I}{\mrm{I}}
\renewcommand{\P}{\mrm{P}}

\newcommand{\Xe}{\mrm{X}}

\renewcommand{\d}{\mrm{d}}
\newcommand{\Cbrcv}{\mrm{C}_\mrm{b}^{\mrm{rcv}}}
\newcommand{\Cbrcp}{\mrm{C}_\mrm{b}^{\mrm{rcp}}}
\newcommand{\Hbf}{\mrm{H}_\mrm{in}}
\newcommand{\Xbank}{\mrm{X}_\mrm{bank}}

\DeclareMathOperator*{\argmin}{arg\,min}

\DeclareMathSymbol{\shortminus}{\mathbin}{AMSa}{"39}

\newcommand{\MPC}{{\scriptscriptstyle\mathrm{MPC}}}
\newcommand{\BC}{{\scriptscriptstyle\mathrm{BC}}}

\usepackage{subcaption} 
\usepackage[table]{xcolor}

\definecolor{bestgreen}{RGB}{198, 230, 201} 
\definecolor{secondblue}{RGB}{187, 222, 251} 
\definecolor{lightgray}{RGB}{210, 210, 210} 

\definecolor{couleur1}{HTML}{0f4c5c}
\definecolor{couleur2}{HTML}{fb8b24}
\definecolor{couleur3}{HTML}{5f0f40}

\definecolor{yellow}{rgb}{1,1,0}
\definecolor{lightgreen}{rgb}{0.5,1,0.5}
\definecolor{darkgray}{rgb}{0.15,0.15,0.15}   
\definecolor{lightgray}{rgb}{0.94,0.94,0.94}  
\definecolor{lightlightgray}{rgb}{0.97,0.97,0.97}  
\definecolor{lightlightblue}{rgb}{0.95,0.97,1.00}  
\definecolor{darkred}{rgb}{0.80,0.00,0.00}    
\definecolor{darkgreen}{rgb}{0.00,0.70,0.00}    
\definecolor{darkblue}{rgb}{0.00,0.00,0.70}    
\definecolor{violet}{rgb}{0.35,0.00,0.70}

\definecolor{lightblue}{cmyk}{0.25,0.15,0,0}
\definecolor{lightred}{cmyk}{0,0.25,0.15,0}
\definecolor{lightgreen}{cmyk}{0.15,0,0.25,0}
\definecolor{lightyellow}{cmyk}{0,0,0.25,0}

\definecolor{darkyellow}{rgb}{0.9,0.9,0.00}    
\definecolor{orange}{rgb}{1.00,0.50,0.00}    
\usepackage{booktabs}

\title{Structured Representation Learning for Behavior Cloning: 
How can we learn to safely control a nuclear power plant?}

\author{%
  Perceval Beja-Battais\textsuperscript{1,2} \quad
  Alain Grossetête\textsuperscript{2} \quad
  Nicolas Vayatis\textsuperscript{1} \\
  \textsuperscript{1}Universit\'e Paris-Saclay, CNRS, ENS Paris-Saclay, Centre Borelli, Gif-sur-Yvette, France \\
  \textsuperscript{2}Framatome, Reactor Process Department, Courbevoie, France \\
  \texttt{perceval.beja-battais@ens-paris-saclay.fr} \\
}

\begin{document}

\maketitle

\begin{abstract}
Learned models for industrial control are usually judged by aggregate accuracy, but accuracy
at the component level does not guarantee safety once it is embedded in the system it is
meant to serve. 
We study this gap on a behavior-cloning task: imitating an expert Nonlinear Model
Predictive Control (NMPC) policy for load-following of a Pressurized Water Reactor (PWR), an 
industrial system with tight safety constraints. 
We propose a structured architecture encoding variables from each timescale into separate latent spaces, 
reflecting the physical decomposition of the system, before training a controller to imitate the expert on the product latent space.
On long-horizon rollouts, separated embeddings improve both accuracy and feasibility compared with a shared-embedding baseline.
Sensitivity analysis further shows that our model yields interpretable
representations aligned with the system's physics. 
However, standalone deployment still leaves several percent of trajectories infeasible regardless of the architecture.
Using our method to warmstart the NMPC optimizer rather than acting standalone, we recover full feasibility and near-optimal cost while still cutting
computation time by $\sim$15\% relative to the expert controller, and even more for abrupt operating changes.
\end{abstract}

\section{Introduction}

\paragraph{Power-grid management and load-following.}
Due to climate change, real-time power grid management increasingly requires Nuclear Power Plants (NPPs) to operate in load-following mode.
To do so, operators can resort to predictive systems~\citep{dupre2025oaps}.
However, operating in load-following can be a challenging task: most NPPs are designed to operate in base-load.

\paragraph{ML for NMPC.}
NMPC is a methodology~\citep{qin2003survey} that repeatedly solves an Optimal Control Problem (OCP) 
minimizing a cost function subject to dynamics and constraints, but its significant online computational burden 
limits its practical deployment in resource-constrained industrial pipelines~\citep{rao2009survey, diehl2009efficient, fiedler2023mpc}.
Machine Learning (ML) has emerged as a principled means of reducing this online cost, 
giving rise to the Learning-based MPC literature~\citep{hewing2020learning}, with methodologies aiming to learn 
either the dynamics~\citep{hossain2023machine} or the controller of a system~\citep{vaupel2020accelerating, li2024faster}. 

\paragraph{Implications for power-grid systems.}
Designing ML-based controllers for multiscale dynamical systems naturally arises for power-grid systems, in which 
the controllers are typically embedded inside a larger system for power-grid management.
Feasibility is crucial: real-time controllers have to provide outputs that strictly respect safety constraints 
inside the nuclear reactor core.

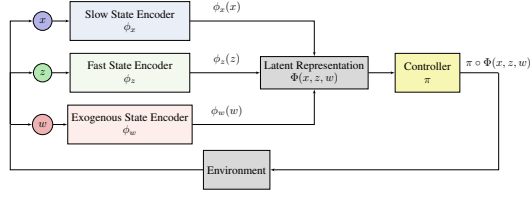
\begin{wrapfigure}{r}{0.6\textwidth}
\centering
\resizebox{0.5\textwidth}{!}{
\begin{tikzpicture}[auto, node distance=2.7cm, >=latex']

    \tikzstyle{block} = [draw, rectangle, minimum height=1.2cm, minimum width=2cm]
    \tikzstyle{block_vert} = [draw, rectangle, minimum height=1cm, minimum width=0.4cm]
    \tikzstyle{sum} = [draw, circle, inner sep=1pt]
    
    \node[circle, draw=black, fill=darkblue!30] (x) at (0,1.6) {$x$};
    \node[circle, draw=black, fill=darkgreen!30] (z) at (0,0) {$z$};
    \node[circle, draw=black, fill=darkred!30] (w) at (0,-1.6) {$w$};
    
    \node[draw, rectangle, right of=x, fill=lightblue!30, align=center, minimum height=1.2cm, minimum width=3.7cm] (B1) {Slow State Encoder\\$\phi_x$};
    \node[right of=B1, node distance=3cm, label=$\phi_x(x)$] (z_x) {};

    \node[draw, rectangle, right of=z, fill=lightgreen!30, align=center, minimum height=1.2cm, minimum width=3.7cm] (B2) {Fast State Encoder\\$\phi_z$};
    \node[right of=B2, node distance=3cm, label=$\phi_z(z)$] (z_z) {};

    \node[draw, rectangle, right of=w, fill=lightred!30, align=center, minimum height=1.2cm, minimum width=3.7cm] (B3) {Exogenous State Encoder\\$\phi_w$};
    \node[right of=B3, node distance=3cm, label=$\phi_w(w)$] (z_w) {};

    \node[block_vert, right of=z_z, fill=gray!30, align=center] (concat) {Latent Representation\\$\Phi(x, z, w)$};
    
    \node[block, right of=concat, node distance=3.5cm, fill=darkyellow!30, align=center] (B4) {Controller\\$\pi$};
    \node[right of=B4, node distance=2.2cm, inner sep=0pt, minimum size=0pt, label={$\pi\circ\Phi(x, z, w)$}] (u) {};
    
    \node[draw=none, inner sep=0pt, minimum size=0pt] (block_before) at (-1,-1.6) {};

    \node[block, fill=gray!30] (env) at (6,-3) {Environment};

    \draw[->] (x) -- (B1.west);
    \draw[->] (z) -- (B2.west);
    \draw[->] (w) -- (B3.west);
    
    \draw[->] (B1) -| (concat.north);
    \draw[->] (B2) -- (concat.west);
    \draw[->] (B3) -| (concat.south);
    \draw[->] (concat.east) -- (B4.west);
    \draw[-] (B4) -- (u);

    \draw[->] (u) -- +(0, -3) -- (env);
    \draw[-] (env) -- +(-7, 0) -- (block_before);
    \draw[->] (block_before) -- +(0, 3.2) -- (x);
    \draw[->] (block_before) -- +(0, 1.6) -- (z);
    \draw[->] (block_before) -- (w);

\end{tikzpicture}%
}
\caption{Proposed multi-stream architecture: we encode the input data in separated latent spaces 
and the controller $\pi$ acts on the product latent space. 
This architecture enhances feasibility of the standalone controller, as well as improving
interpretability and closed-loop performance.}
\label{fig:summarize_controller}
\end{wrapfigure}

\paragraph{Contributions.}
We propose Structured Representation Learning for 
Behavior Cloning (SRLBC) of an expert NMPC controller for a nuclear reactor core.
We design a structured closed-loop controller and show that
separating timescales yields more interpretable representations,
improves performance and the satisfaction of strict nonlinear safety constraints.
Moreover, we show that using the learned policy to warmstart the expert NMPC optimizer recovers full constraint feasibility while still yielding a significant reduction 
in computation time, offering a safety-preserving deployment path for the proposed controller.

\section{Problem Statement}\label{sec:PbStatement}

\paragraph{Dynamical system and control problem.}
The goal is to design a closed-loop controller for the load-following of PWRs.
The dynamics of the core evolve according to the differential system
\begin{equation}\label{eq:full_dynamics}
    \begin{aligned}
        \dot{x}(t) &= f(x(t), z(t), u(t), w(t)) \in \R^{n_x}, \quad x(0) = x_0,\\
        \varepsilon\dot{z}(t) &= g(x(t), z(t), u(t), w(t)) \in \R^{n_z}, \quad z(0) = z_0. \\
    \end{aligned}
\end{equation}
The system is described by a slow state variable $x(t)$, a fast state variable $z(t)$, and an exogenous input variable $w(t)$ which is known in advance (e.g., 
power demand).
The full equations are given in Appendix~\ref{appendix}. 
$u(t)$ represents dilution or boron injection commands from the operator. 
This system is highly nonlinear due to multiple factors: stiffness arises from 
the fast-evolving coupled subsystem on $z(t)$, with challenging nonlinearities 
induced by the reactivity (a highly nonlinear function, depending on all the state variables of the system~\citep{stacey2018nuclear}).
Moreover, $u(t)$ directly acts on the slow state $x(t)$, but differently depending on its sign: 
$\frac{\partial f}{\partial u}$ is discontinuous at $u = 0$.
The substantially high dimension of the control problem, with $n_x = 14, n_z = 8, n_w = 30$,
induces a non-trivial mapping $(x(t), z(t), w(t), \dots, w(t+k \ \d t)) \mapsto \pi_\MPC(x,z,w) \coloneqq (u^*(t), \dots, u^*(t+k \ \d t))$. 
Our aim is to design an explicit state-feedback controller, without online optimization,
mimicking the behavior of the optimal control input $u^*(t)$ of the nonlinear OCP
\begin{equation}\label{eq:OCP}
        u^*\!\coloneqq\!\argmin_{u}\!\left[J(u)\!\coloneqq\!\int_{0}^{t_f}\!\ell(x(t), z(t), w(t), u(t)) \d t\right],
        \quad \text{s.t.~$h(z(t)) \leq 0$ and~\eqref{eq:full_dynamics}.}
\end{equation}

\paragraph{From the OCP to Behavior Cloning.}
Let $\pi_\MPC(x_0, z_0, w)$ be the optimal policy recovered from solving~\eqref{eq:OCP} at each timestep. 
Approximate MPC (AMPC) learns an ML-based policy that can be deployed online without solving the underlying OCP, while providing statistical guarantees~\citep{hertneck2018learning}. 
However, the considered robust MPC formulation relies on a quadratic cost function in the state and control variables, which limits its applicability to our application setting.
Imitation Learning (IL) consists of a wider class of methods, in which a policy $\pi$ is learned by observing an expert's demonstration~\citep{hua2021learning}. 
The simplest IL framework is Behavior Cloning (BC).
In this setting, we assume access to a dataset of optimal control inputs and trajectories $\mathcal{D} \coloneqq \{(x_i, z_i, w_i, \pi_{\MPC}(x_i, z_i, w_i)), i\in \llbracket 1, N\rrbracket\}$, 
and learn a mapping $\pi_\BC \approx \pi_\MPC$.
Learning such policies for complex systems can be challenging, especially when the offline computational cost required to generate a representative 
set of demonstrations from $\pi_\MPC(x,z,w)$ is substantial. 
Moreover, statistical learning theory is difficult to apply to provide safety guarantees for the resulting controller.
In most cases, $\pi_\BC$ may still act as a black box. 
For safety-critical applications, $\pi_\BC$ should therefore be designed to be as interpretable as possible, even when used for warmstarting. 

\section{Experiment} \label{sec:Model}

\paragraph{Designing the controller and training the model.}
Our model consists of four components:
a slow-state encoder $\phi_x : \mathbb{R}^{n_x} \rightarrow \mathbb{R}^{d_x}$,
a fast-state encoder $\phi_z : \mathbb{R}^{n_z} \rightarrow \mathbb{R}^{d_z}$,
an exogenous-variable encoder $\phi_w : \mathbb{R}^{n_w} \rightarrow \mathbb{R}^{d_w}$,
and a latent controller
$\pi_{\BC} : \mathbb{R}^{d_x+d_z+d_w} \rightarrow \mathbb{R}^{n_u}$.
The entire architecture is trained via BC to approximate the expert controller
$\pi_{\MPC}(x,z,w)$ by minimizing the BC loss $\ell_{\BC}(\cdot)
    \coloneqq
    \left\|
    \pi_{\BC}\circ\Phi(\cdot)
    -
    \pi_{\MPC}(\cdot)
    \right\|$,
where $\Phi \coloneqq [\phi_x^\top, \phi_z^\top, \phi_w^\top]^\top$.
We used a 2-layer MLP with $d = 64$ neurons per hidden layer and ReLU activation for each encoder $\phi_x(x), \phi_z(z), \phi_w(w)$.
The controller block has the same architecture.
We give further motivations for these choices in Appendix~\ref{appendix:prior}.
The encoders were pretrained on the reconstruction loss, after which the main training consists of 
simultaneously optimizing the encoders and controllers on the BC loss in order 
to finetune the encoders specifically for control tasks.
Compared with a naive shared embedding, this architectural prior imposes a strong structural constraint on the representation.
The dimensions of each embedding were selected by comparing a validation BC loss over all 
possible 91 combinations in $\{(d_x, d_z, d_w), \ n_x \leq d_x, n_z \leq d_z, n_w \leq d_w, \ d = d_x + d_z + d_w\}$, 
so that each embedding is at least as large as the underlying dimension.
Note that this latent-dimension search is specific to the separated architecture, 
which must allocate capacity across three constrained subspaces a priori. 
We compare the performance of the proposed methodology with an unconstrained embedding of dimension $d$ on $(x, z, w)$.
For the common baseline, no such decomposition needs to be chosen.
With this methodology, we found $d_x = 14, \ d_z = 12, \ d_w = 38$.
Common embeddings and separated embeddings share the same total latent 
dimension $d = 64 = d_x + d_z + d_w$, ensuring a fair comparison.
We also compared our methodology against a larger, encoder-controller baseline, but with $2d = 128$ neurons per layer, 
and a $2d$-dimensional embedding.
Finally, we compare SRLBC with and without enforced decorrelation. 
This is done by also minimizing $\lambda\|\mathrm{Cov}(\phi_\alpha) - I_{d_\alpha}\|, \ \alpha \in \{x, z, w\}$ during training,
with $\lambda$ a hyperparameter that was optimized using log-grid search.

\definecolor{oursblue}{RGB}{230,240,250}
\definecolor{bestgreen}{RGB}{220,245,220}
\definecolor{secondyellow}{RGB}{255,245,210}

\begin{table*}[t]
\centering
\renewcommand{\arraystretch}{1.25}
\resizebox{\textwidth}{!}{
{\small
\begin{tabular}{|c|cccc||ccc|}
\specialrule{1.2pt}{0pt}{0pt}

\textbf{Method} & 
\multicolumn{4}{c||}{\textbf{Without Warmstart (Standalone)}}
&
\multicolumn{3}{c|}{\textbf{With Warmstart}}
\\

\specialrule{0.8pt}{0pt}{0pt}

\textbf{Embed. (Size) \& Train. Spec.}
&
\textbf{Ratio to Optimal Cost}
&
\textbf{Relative Gap to Optimal Control}
&
\textbf{Feasibility}
&
\textbf{Comput. Time (s)}
&
\textbf{Ratio to Optimal Cost}
&
\textbf{Feasibility}
&
\textbf{Comput. Time (s)}
\\

\specialrule{1.2pt}{0pt}{0pt}

BC Baseline (64)
& $1.12 \pm 0.12$
& $(8.0 \pm 2.8)\cdot 10^{-3}$
& $94.1\%$
& $20.4\pm0.6$
& \cellcolor{bestgreen}$\mathbf{1.00 \pm 0.01}$
& \cellcolor{bestgreen}$\mathbf{100.0\%}$
& $292.3\pm 17.9$
\\

BC Baseline (128)
& $1.15 \pm 0.15$
& $(9.8 \pm 3.9)\cdot 10^{-3}$
& $92.2\%$
& $20.5\pm0.6$
& \cellcolor{bestgreen}$\mathbf{1.00 \pm 0.01}$
& $99.6\%$
& $293.8\pm 17.6$
\\

\specialrule{0.8pt}{0pt}{0pt}

SRLBC (64)
& $1.08 \pm 0.11$
& $(7.8 \pm 3.5)\cdot 10^{-3}$
& \cellcolor{bestgreen}$\mathbf{97.8\%}$
& $20.5\pm0.6$
& \cellcolor{bestgreen}$\mathbf{1.00 \pm 0.01}$
& \cellcolor{bestgreen}$\mathbf{100.0\%}$
& \cellcolor{bestgreen}$\mathbf{285.7\pm 16.1}$
\\

SRLBC (64) + Decor.
& \cellcolor{bestgreen}$\mathbf{1.07 \pm 0.08}$
& \cellcolor{bestgreen}$\mathbf{(6.8 \pm 2.5)\cdot 10^{-3}}$
& $95.3\%$
& $20.5\pm0.6$
& \cellcolor{bestgreen}$\mathbf{1.00 \pm 0.01}$
& \cellcolor{bestgreen}$\mathbf{100.0\%}$
& $293.0\pm 17.3$
\\

\specialrule{0.8pt}{0pt}{0pt}

\rowcolor{gray!10}
Optimal Policy -- $\pi_{\MPC}$
& $\mathit{1.00}$
& $\mathit{0}$
& $\mathit{100.0\%}$
& $\mathit{348.3\pm14.6}$
& $\mathit{1.00}$
& $\mathit{100.0\%}$
& $\mathit{348.3\pm14.6}$
\\

\specialrule{1.2pt}{0pt}{0pt}
\end{tabular}
}
}

\caption{Comparison of the learned policies in standalone operation 
and when used to warmstart the NMPC optimization. 
Cost ratio to optimal $=J(\pi_{\BC}\circ\Phi)/J(\pi_{\MPC})$, 
Rel. gap to optimal control $=\|\pi_{\BC}\circ\Phi-\pi_{\MPC}\|/\|\pi_{\MPC}\|$.
$\mathrm{Mean}\pm\mathrm{Std}$ over 1000 test rollouts of 48h. 
Without warmstart, $\pi_{\BC}\circ\Phi$ is directly used as the control policy. 
With warmstart, the policy provides the initial guess for the NMPC optimization at each timestep.}
\label{tab:quantitative_results}
\end{table*}

\paragraph{Data acquisition.} \label{subsec:Data}
To design the dataset $\mathcal{D} \coloneqq \{(x_i, z_i, w_i, \pi_{\MPC}(x_i, z_i, w_i)),~i\in \llbracket 1, N\rrbracket\}$, 
a large number of simulations were collected in which the expert NMPC controller optimized the nonlinear OCP~\eqref{eq:OCP}. 
The stage cost of the integral cost function $\ell(z(t), w(t))$ penalizes the distance of the axial offset of the power distribution of the core from a reference.  
Moreover, a strict safety constraint $h(z(t)) \leq 0$ on the axial offset deviation has to be respected at all times. 
This OCP involves a non-convex, nonlinear cost function, non-convex nonlinear constraints over 
the complex multiscale system~\eqref{eq:full_dynamics}.
The simulations were sampled using a random transient generator, which corresponds to a realistic 
random exogenous sequence of $w(t)$, e.g., low demand at night, high demand in mornings and evenings.
In total, approximately $80\,000$ training points of the form 
$(x_i, z_i, w_i, \pi_{\MPC}(x_i, z_i, w_i))$ were collected through parallel simulations of $260$ trajectories, each spanning a duration of $48$ hours. 
Using DAgger~\citep{ross2011reduction} for each benchmarked policy, the training set was augmented with $100$ new $48$-hour trajectories.

\paragraph{Quantitative Results.}
We compared the performance of the different architectures on a large test set of $1\,000$ 
rollouts of 48h in Table~\ref{tab:quantitative_results}.
Using SRLBC, the controller learns a mapping closer to $\pi_\MPC$, indicating 
that separated embeddings help the learned policy to better approximate the expert controller.
Rollouts using the NMPC controller yield the optimal trajectory 
in about 6 minutes for 48h simulations, while rollouts using any BC-based controller are rapid due to their instantaneous online evaluation.
In practice, 20 seconds roughly corresponds to the computation 
time required for simulating the system during 48h.
This yields an acceleration of a factor of $\sim$17 compared to the expert when BC controllers act standalone.
However, we recommend combining the BC approach with a form of safety-filter as is often done for sensitive applications~\citep{hsu2023safety}. 
In our case, this can be done by warmstarting the optimization process of the expert controller with the outputs of the learned controller. 
This allows systematic feasibility recovery, while still yielding an acceleration of about 15\% on average.
More specifically, the speedup is particularly significant when standard shift-initialization fails (implemented on $\pi_\MPC$). 
If $w(t)$ varies over the last segment of the control discretization, e.g., due to changes in power demand, 
BC-based methods are substantially more efficient.
When the electricity regulator asks for a last-minute change, these gains really matter.

\paragraph{Interpretability of the controller.}
\begin{wrapfigure}{r}{0.3\textwidth}
    \centering
    \begin{subfigure}{0.9\linewidth}
        \centering
        \includegraphics[width=\linewidth]{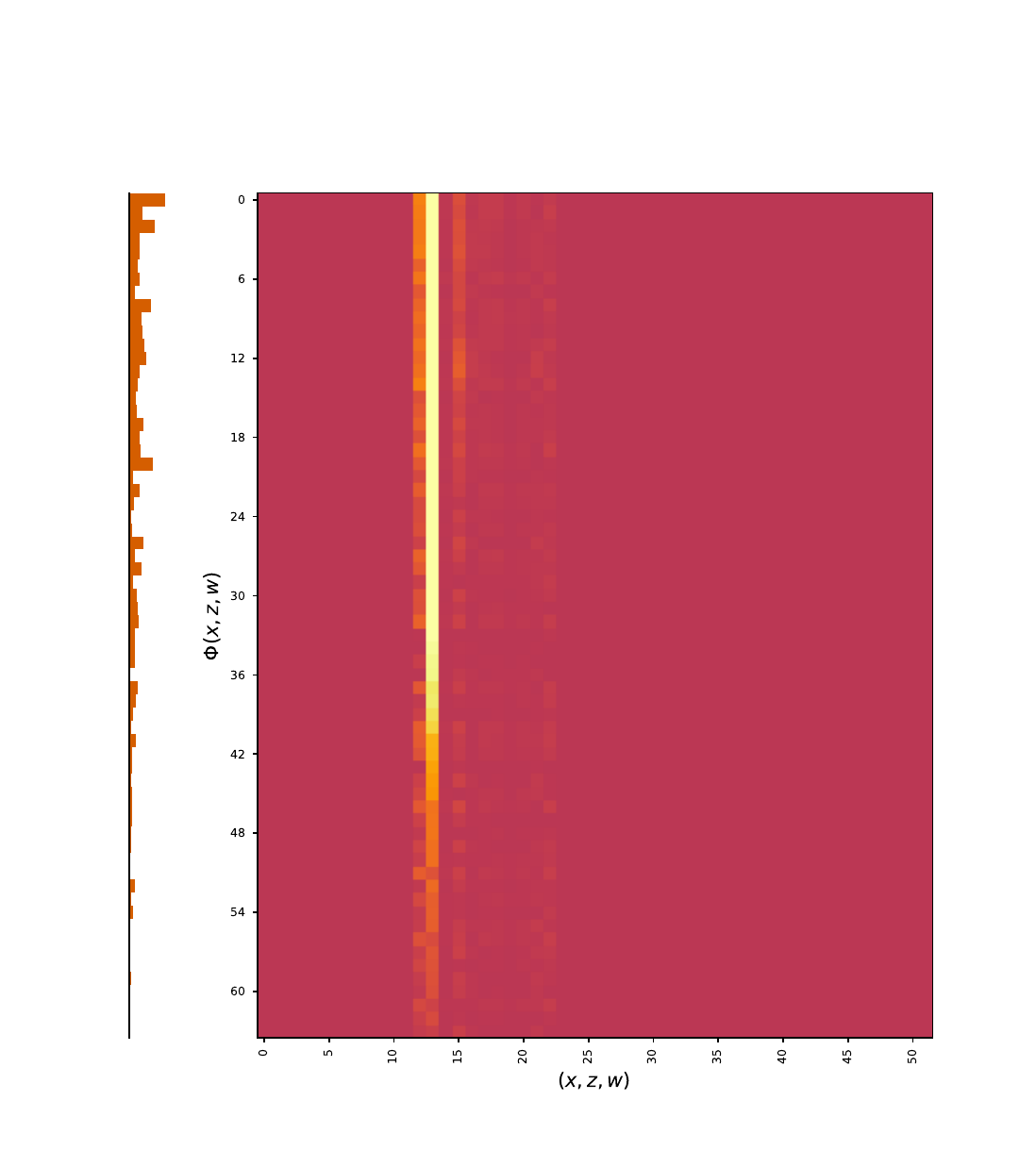}
        \caption{Unstructured embedding.}
        \label{fig:sensitivity_analysis_1}
    \end{subfigure}

    \vspace{0.5em}

    \begin{subfigure}{0.9\linewidth}
        \centering
        \includegraphics[width=\linewidth]{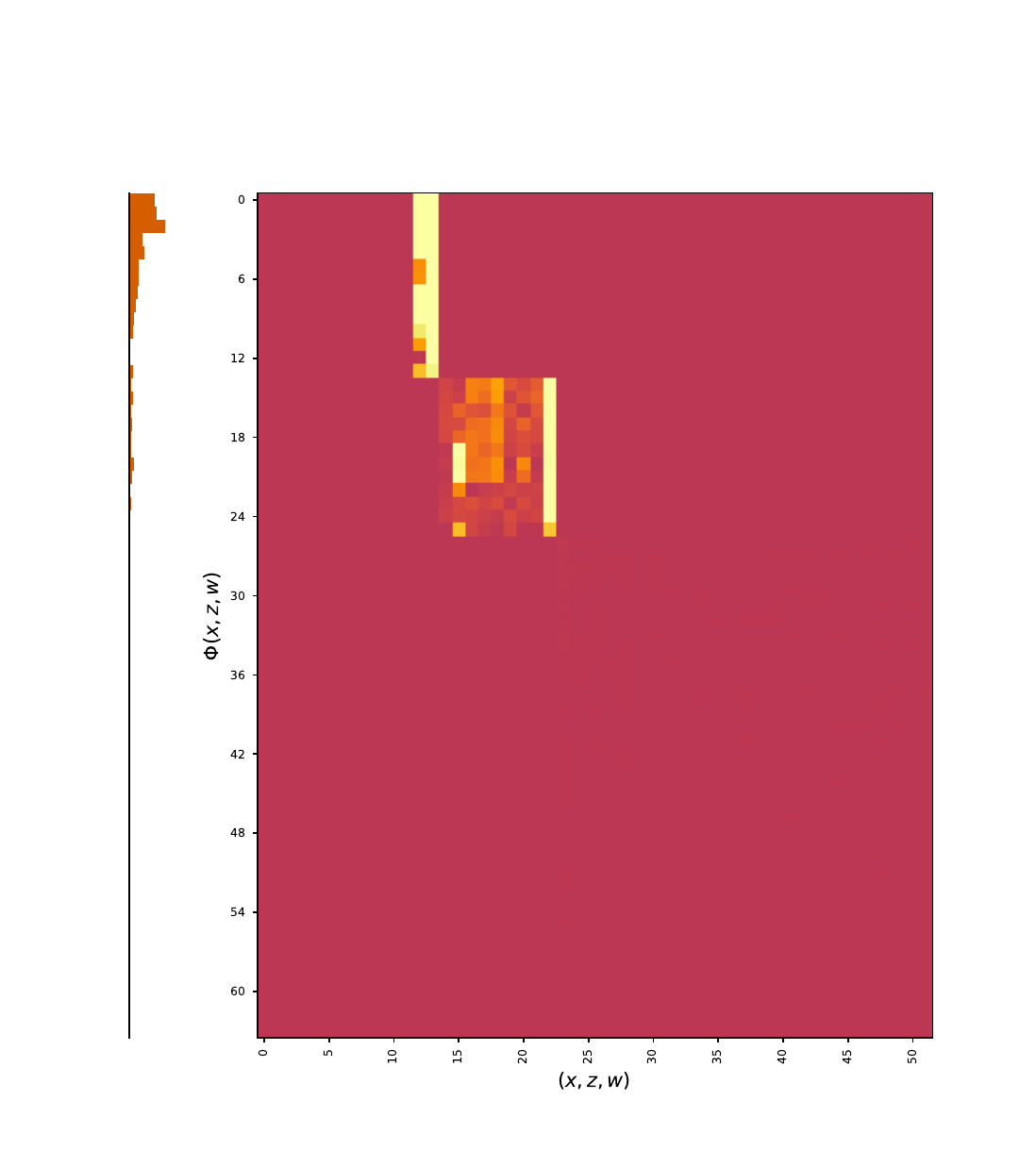}
        \caption{Embedding from SRLBC.}
        \label{fig:sensitivity_analysis_2}
    \end{subfigure}
    \caption{Sensitivity analysis.}
\end{wrapfigure}
The separated architecture imposes a block-structured decomposition of the latent space.
This yields a more interpretable embedding, 
directly depending on understandable combinations of physical variables.
The heatmaps from Figs.~\ref{fig:sensitivity_analysis_1}--\ref{fig:sensitivity_analysis_2} 
show the averaged absolute value of the Jacobian coefficients for each embedding (normalized
with respect to the input's variances).
The histograms, on the left side, show relative importance of the $i^\mathrm{th}$ representation $[\Phi(x, z, w)]_i$
for constructing the controller's output $\pi \circ \Phi(x, z, w)$.
We recover that a few state variables (namely, the boron concentration and the fast-evolving variables from $z(t)$) explain most 
of the controller's output.
The rest of the state and exogenous variables help regularize.
This observation is consistent with reactor physics, since the control input directly impacts the boron 
concentration, and because the cost and constraint functions only depend on $z(t)$. 
What is most striking, however, is that separating the embeddings of the slow and fast variables 
drastically improves interpretability: $\frac{\partial \pi}{\partial \Phi}$ is much more structured 
and less diffuse throughout the components.

\section{Discussion}\label{sec:Discussion}

\paragraph{Feasibility is not a byproduct of accuracy.}
A recurring assumption in learning-based control is that minimizing an imitation loss is
a good proxy for safe deployment. Our results suggest otherwise. 
The BC loss and the closed-loop cost ratio track each other closely across all architectures, yet
feasibility varies by up to 6 points among architectures with similar losses.
The decoupling is, we believe, a useful cautionary example for the broader question of what it means 
for a learned model to \emph{respect the physics}: an aggregate error metric can improve while 
the property that actually matters for deployment moves in the opposite direction. 

\paragraph{Components are not systems.}
Table~\ref{tab:quantitative_results} tells two different stories about the same controller. 
Standalone, even our best BC variants leave 2 to 5\% of trajectories infeasible but are not, on their own, safe
substitutes for the loop they would replace. 
Used instead as a warmstart for the expert NMPC optimizer,
the same policies recover full feasibility while still cutting computation time. 
Here, the controller did not change, but its role in the system did. 

\paragraph{Limitations.}
Two limitations bound the scope of these claims. 
First, the benchmark exhibits distributional shift within a fixed system (varying xenon dynamics, 
exogenous turbine-power profiles) rather than a structural shift. 
Second, the benchmark itself is a single-plant industrial simulator, not an open community dataset, 
and is closer in spirit to the kind of high-fidelity, non-IEEE test case this workshop calls for than to a released artifact.


\section{Conclusion}

We introduced SRLBC, a simple timescale-aware architecture for behavior-cloning of expert controllers of
multiscale nonlinear systems.
We tried to answer the larger question: what should we require from a learned controller before using it as an embedded system?
On a challenging PWR load-following benchmark, we find that BC metrics can be poor proxies for constraint satisfaction:
accuracy alone does not predict the controller's effect once embedded in a system.
By itself, structuring the embedding by timescale improves feasibility and interpretability. 
Combining it with NMPC-warmstarting recovers full constraint satisfaction
with a significant compute saving over the expert controller, yielding an 
improved robust controller. 

\bibliographystyle{plainnat}
\bibliography{refs} 

\begin{thebibliography}{18}
\providecommand{\natexlab}[1]{#1}
\providecommand{\url}[1]{\texttt{#1}}
\expandafter\ifx\csname urlstyle\endcsname\relax
  \providecommand{\doi}[1]{doi: #1}\else
  \providecommand{\doi}{doi: \begingroup \urlstyle{rm}\Url}\fi

\bibitem[Andersson et~al.(2019)Andersson, Gillis, Horn, Rawlings, and
  Diehl]{andersson2019casadi}
Joel~AE Andersson, Joris Gillis, Greg Horn, James~B Rawlings, and Moritz Diehl.
\newblock {CasADi}: a software framework for nonlinear optimization and optimal
  control.
\newblock \emph{Mathematical Programming Computation}, 11\penalty0
  (1):\penalty0 1--36, 2019.

\bibitem[Diehl et~al.(2009)Diehl, Ferreau, and Haverbeke]{diehl2009efficient}
Moritz Diehl, Hans~Joachim Ferreau, and Niels Haverbeke.
\newblock Efficient numerical methods for nonlinear {MPC} and moving horizon
  estimation.
\newblock In \emph{Nonlinear model predictive control}, pages 391--417.
  Springer, 2009.

\bibitem[Dupr{\'e} and Grosset{\^e}te(2025)]{dupre2025oaps}
Guillaume Dupr{\'e} and Alain Grosset{\^e}te.
\newblock The {OAPS} solution: a real-time predictive system for flexible {PWR}
  operation.
\newblock In \emph{International Congress on Advances in Nuclear Power Plants
  (ICAPP)}, 2025.

\bibitem[Dupr{\'e} et~al.(2021)Dupr{\'e}, Grosset{\^e}te, Chevrel, and
  Yagoubi]{dupre2021enhanced}
Guillaume Dupr{\'e}, Alain Grosset{\^e}te, Philippe Chevrel, and Mohamed
  Yagoubi.
\newblock Enhanced flexibility of {PWR}s ({M}ode {A}) using an efficient
  {NMPC}-based boration/dilution system.
\newblock In \emph{2021 european control conference (ECC)}, pages 1092--1098.
  IEEE, 2021.

\bibitem[Fiedler et~al.(2023)Fiedler, Karg, L{\"u}ken, Brandner, Heinlein,
  Brabender, and Lucia]{fiedler2023mpc}
Felix Fiedler, Benjamin Karg, Lukas L{\"u}ken, Dean Brandner, Moritz Heinlein,
  Felix Brabender, and Sergio Lucia.
\newblock {do-mpc}: Towards {FAIR} nonlinear and robust model predictive
  control.
\newblock \emph{Control Engineering Practice}, 140:\penalty0 105676, 2023.

\bibitem[Hertneck et~al.(2018)Hertneck, K{\"o}hler, Trimpe, and
  Allg{\"o}wer]{hertneck2018learning}
Michael Hertneck, Johannes K{\"o}hler, Sebastian Trimpe, and Frank
  Allg{\"o}wer.
\newblock Learning an approximate model predictive controller with guarantees.
\newblock \emph{IEEE Control Systems Letters}, 2\penalty0 (3):\penalty0
  543--548, 2018.

\bibitem[Hewing et~al.(2020)Hewing, Wabersich, Menner, and
  Zeilinger]{hewing2020learning}
Lukas Hewing, Kim~P Wabersich, Marcel Menner, and Melanie~N Zeilinger.
\newblock Learning-based model predictive control: Toward safe learning in
  control.
\newblock \emph{Annual Review of Control, Robotics, and Autonomous Systems},
  3\penalty0 (1):\penalty0 269--296, 2020.

\bibitem[Hossain and Kumar(2023)]{hossain2023machine}
Ramij~Raja Hossain and Ratnesh Kumar.
\newblock Machine learning accelerated real-time model predictive control for
  power systems.
\newblock \emph{IEEE/CAA Journal of Automatica Sinica}, 10\penalty0
  (4):\penalty0 916--930, 2023.

\bibitem[Hsu et~al.(2023)Hsu, Hu, and Fisac]{hsu2023safety}
Kai-Chieh Hsu, Haimin Hu, and Jaime~F Fisac.
\newblock The safety filter: A unified view of safety-critical control in
  autonomous systems.
\newblock \emph{Annual Review of Control, Robotics, and Autonomous Systems}, 7,
  2023.

\bibitem[Hua et~al.(2021)Hua, Zeng, Li, and Ju]{hua2021learning}
Jiang Hua, Liangcai Zeng, Gongfa Li, and Zhaojie Ju.
\newblock Learning for a robot: Deep reinforcement learning, imitation
  learning, transfer learning.
\newblock \emph{Sensors}, 21\penalty0 (4):\penalty0 1278, 2021.

\bibitem[Li et~al.(2024)Li, Wang, Chen, Paleja, Nageshrao, and
  Gombolay]{li2024faster}
Zhaoxin Li, Xiaoke Wang, Letian Chen, Rohan Paleja, Subramanya Nageshrao, and
  Matthew Gombolay.
\newblock Faster model predictive control via self-supervised initialization
  learning.
\newblock \emph{arXiv preprint arXiv:2408.03394}, 2024.

\bibitem[Paszke et~al.(2019)Paszke, Gross, Massa, Lerer, Bradbury, Chanan,
  Killeen, Lin, Gimelshein, Antiga, et~al.]{paszke2019pytorch}
Adam Paszke, Sam Gross, Francisco Massa, Adam Lerer, James Bradbury, Gregory
  Chanan, Trevor Killeen, Zeming Lin, Natalia Gimelshein, Luca Antiga, et~al.
\newblock {PyTorch}: An imperative style, high-performance deep learning
  library.
\newblock \emph{Advances in neural information processing systems}, 32, 2019.

\bibitem[Qin and Badgwell(2003)]{qin2003survey}
S~Joe Qin and Thomas~A Badgwell.
\newblock A survey of industrial model predictive control technology.
\newblock \emph{Control engineering practice}, 11\penalty0 (7):\penalty0
  733--764, 2003.

\bibitem[Rao(2009)]{rao2009survey}
Anil~V Rao.
\newblock A survey of numerical methods for optimal control.
\newblock \emph{Advances in the Astronautical Sciences}, 135\penalty0
  (1):\penalty0 497--528, 2009.

\bibitem[Ross et~al.(2011)Ross, Gordon, and Bagnell]{ross2011reduction}
St{\'e}phane Ross, Geoffrey Gordon, and Drew Bagnell.
\newblock A reduction of imitation learning and structured prediction to
  no-regret online learning.
\newblock In \emph{Proceedings of the fourteenth international conference on
  artificial intelligence and statistics}, pages 627--635. JMLR Workshop and
  Conference Proceedings, 2011.

\bibitem[Stacey(2018)]{stacey2018nuclear}
Weston~M Stacey.
\newblock \emph{Nuclear reactor physics}.
\newblock John Wiley \& Sons, 3rd edition, 2018.

\bibitem[Vaupel et~al.(2020)Vaupel, Hamacher, Caspari, Mhamdi, Kevrekidis, and
  Mitsos]{vaupel2020accelerating}
Yannic Vaupel, Nils~C Hamacher, Adrian Caspari, Adel Mhamdi, Ioannis~G
  Kevrekidis, and Alexander Mitsos.
\newblock Accelerating nonlinear model predictive control through machine
  learning.
\newblock \emph{Journal of process control}, 92:\penalty0 261--270, 2020.

\bibitem[W{\"a}chter and Biegler(2006)]{wachter2006implementation}
Andreas W{\"a}chter and Lorenz~T Biegler.
\newblock On the implementation of an interior-point filter line-search
  algorithm for large-scale nonlinear programming.
\newblock \emph{Mathematical programming}, 106\penalty0 (1):\penalty0 25--57,
  2006.

\end{thebibliography}

\appendix
\section{Reactor Core Dynamics}\label{appendix}

The system's dynamics are given by:
\begin{subequations}\label{eq:dynmical_system}
\par\noindent
\begin{minipage}[t]{0.48\textwidth}
    \small
    \begin{align}
        \frac{\d \I}{\d t}\!&=\!N\gamma_\I \P(t) - \lambda_\I \I(t), \\
        \frac{\d \Xe}{\d t}\!&=\!N\gamma_\Xe \P(t)\!+\!\lambda_\I \I(t) \nonumber \\
        &\quad -\!(\lambda_\Xe\!+\!\sigma_\Xe N \P(t))\Xe(t), \\
        \frac{\d \Cbrcv}{\d t}\!&=\!\frac{1}{\tau_{\mathrm{rcv}}}\left(-\Cbrcv(t) + C_\mathrm{b, inj}(t)\right), \\
        \frac{\d \Cbrcp}{\d t}\!&=\!\frac{Q}{M_{\mathrm{rcp}}}(\Cbrcv(t) - \Cbrcp(t)),
    \end{align}
\end{minipage}%
\hfill
\begin{minipage}[t]{0.51\textwidth}
    \small
    \begin{align}
        \frac{\d \Hbf}{\d t}\!&=\!\frac{K_\mathrm{H}}{\tau_{\mathrm{sg}}}\Big(\textstyle\sum_{i=1}^{N}\P_i(t) - P_\mathrm{turb}(t)\Big), \\
        \frac{\d \Xbank}{\d t}\!&=\!\mathrm{GF4}(T_{\mathrm{ref}}(P_\mathrm{turb}(t)) - T(\Hbf(t))), \\
        \frac{\d \P}{\d t}\!&=\!\left(\frac{(\rho(t) - \beta) I_{N} + D_{\mathrm{exch}}}{\Lambda^*}\right) \P(t),
    \end{align}
\end{minipage}
\par
\end{subequations}
\noindent where $C_\mathrm{b, inj}$ follows the equation
            \begin{equation*}
                    C_\mathrm{b, inj}(t) \coloneqq \left\{\begin{aligned}
                    \Cbrcp(t) + \frac{u(t)}{Q_{\mathrm{ch}}}\left(C_{\mathrm{bor, rea}} - \Cbrcp(t)\right)~\text{if $u(t) \geq 0$,} \\
                    \Cbrcp(t) - \frac{u(t)}{Q_{\mathrm{ch}}}\left(C_{\mathrm{dil, rea}} - \Cbrcp(t)\right)~\text{if $u(t) < 0$.}
                \end{aligned}\right.
            \end{equation*}
$\mathrm{GF4}$ is a discontinuous profile. 
All the other undefined parameters are either scalar or matrix constants.
In these equations, $x(t) \coloneqq (\I(t), \Xe(t), \Cbrcv(t), \Cbrcp(t))$, and $z(t) \coloneqq (\Hbf(t), \Xbank(t), \P(t))$. 
$\I, \Xe$, and $\P$ are 6-dimensional vectors.
One can introduce a ratio $0 < \varepsilon \approx 10^{-2} \ll 1$ between the typical timescales of 
slow and fast variables.

\section{Infeasibility example}\label{appendix:example}

\begin{figure}[H]
    \centering
    \includegraphics[width=0.8\textwidth]{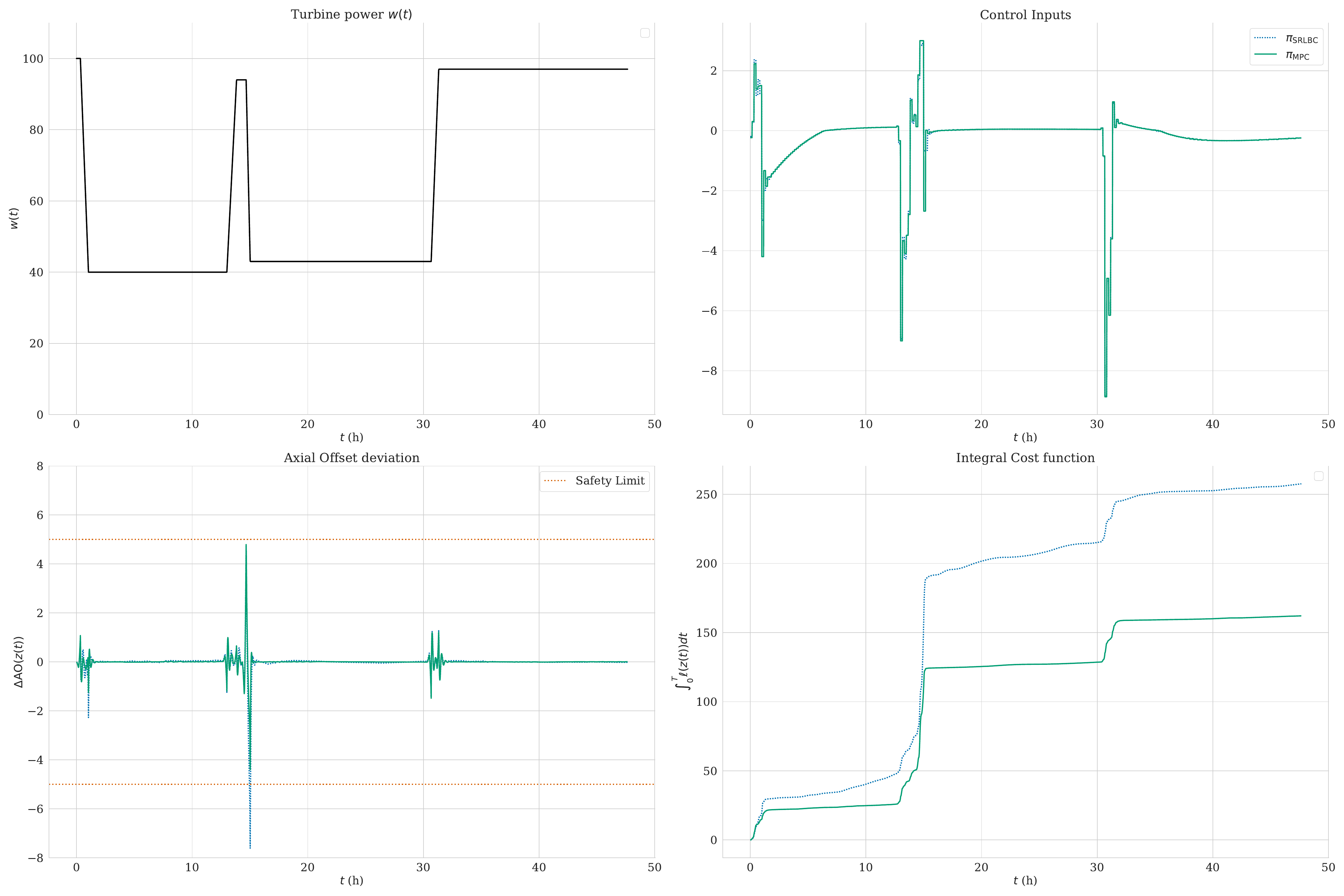}
    \caption{Representative example of a failure using SRLBC standalone controller (dotted, blue). 
    NMPC expert is plotted in solid green.
    At $t\approx 15$ hours, a last-minute change occurs: the power that was increasing is rapidly
    asked to decrease, which creates a challenging situation. 
    SRLBC fails to keep the axial offset above $-5\%$ (bottom-left graph).}
\end{figure}

\section{Details on the formulation, solver and hardware}\label{appendix:details}

\paragraph{NMPC formulation.}
The expert controller $\pi_\MPC$ solves the OCP~\eqref{eq:OCP} in a receding-horizon fashion, following 
the methodology from~\citep{dupre2021enhanced}.
The OCP is solved over a prediction horizon of $K = 30$ steps, 
transcribed using single shooting formulation with $\d t = 1~$min.
The control input is parametrized as piecewise constant over each interval 
$[t + 10j\,\d t, t + 10(j+1)\,\d t)$, $j \in \{0, 1, 2\}$,
which yields the discretized problem
\begin{equation}\label{eq:discrete_OCP}
    \begin{aligned}
        \min_{u = (u_0, u_1, u_2)} \quad & \sum_{j=0}^{2} \sum_{k=0}^{9} \ell(x_{10j+k}, z_{10j+k}, w_{10j+k}, u_j) \\
        \text{s.t.} \quad & (x_{i+1}, z_{i+1}) = F_{\d t}(x_i, z_i, u_j, w_i), \quad (x_0, z_0) = (x(t), z(t)), \\
        & h(z_{i+1}) \leq 0, \quad u_{\min} \leq u_j \leq u_{\max}, \quad i \in \{0, \dots, K-1\}, \quad j = \lfloor i/10 \rfloor,
    \end{aligned}
\end{equation}
where $F_{\d t}$ denotes the numerical integration of the stiff dynamics~\eqref{eq:full_dynamics} over one sampling interval,
performed with a specialized homemade stiff solver.
The stage cost is $\ell(z(t), w(t)) = \Biggl|\underbrace{\frac{\sum_{i=1}^{3} \P_i - \sum_{i=4}^{6} \P_i}{\sum_{i=1}^{6} \P_i}}_{\coloneqq \mathrm{AO}(\P(t))} - \mathrm{AO}_\mathrm{ref}(w(t))\Biggr|^2$,
the safety constraint is $h(z(t)) = |\mathrm{AO}(\P(t)) - \mathrm{AO}_{\mathrm{ref}}(w(t))| - \Delta\mathrm{AO}_{\max}$ with $\Delta\mathrm{AO}_{\max} = 5\%$,
and the input bounds are $u_{\min} = -10~$kg/s and $u_{\max} = 3~$kg/s.

\paragraph{Initialization and warmstart.}
In the baseline expert $\pi_\MPC$, the optimizer is initialized by shift-initialization:
the solution computed at time $t$ is shifted by one step, and the last control input is
duplicated, to provide the initial guess at $t + \d t$.
In the warmstarted configuration, the initial guess is instead given by the output of the learned policy,
$(u_0^{(0)}, u_1^{(0)}, u_2^{(0)}) \coloneqq \pi_\BC \circ \Phi(x(t), z(t), w)$,
which predicts the whole control sequence over the horizon.
The optimizer then proceeds identically in both cases, so that feasibility and optimality are certified by the solver rather than by the learned policy.

\paragraph{Solver.}
The discretized problem~\eqref{eq:discrete_OCP} is solved with IPOPT~\citep{wachter2006implementation} via \texttt{CasADi}~\citep{andersson2019casadi}.
The stopping criteria are the algorithm tolerance, set to $0.1$, and the maximum iteration number set to $30$.  

\paragraph{Hardware and deployment.}
All closed-loop rollouts reported in Table~\ref{tab:quantitative_results}, for both the NMPC expert and the learned controllers,
were run on a single core of an Intel Xeon Platinum 8260 (2.40 GHz).
Both the NMPC and the neural network controllers are deployed on the CPU,
so that the reported computation times are directly comparable.
The neural networks were trained on a single NVIDIA Tesla V100-SXM2-32GB, using \texttt{PyTorch}~\citep{paszke2019pytorch}.
Training one model from scratch takes a few minutes ($\leq 10~$minutes).

\section{Motivation for the block-separable representation}\label{appendix:prior}

\paragraph{Where the coupling is modeled.}
The block-separable structure only constrains the encoding of the variables, not the decision.
Each encoder $\phi_x$, $\phi_z$, $\phi_w$ extracts a representation of a single group of variables,
while the latent controller $\pi_\BC$ is a nonlinear MLP acting on the concatenation $\Phi = [\phi_x^\top, \phi_z^\top, \phi_w^\top]^\top$.
Interactions between $x$, $z$ and $w$ are therefore still
learned by $\pi_\BC$, which can represent arbitrary nonlinear couplings between the three latent blocks.
The prior we impose is that the coupling between variables should be resolved at the controller level,
once each group of variables has been summarized separately, rather than mixed from the first layer.

\paragraph{Why this prior matches the structure of the dynamics.}
The dynamics~\eqref{eq:dynmical_system} are coupled but heavily structured.
First, the boron subsystem $(\Cbrcv, \Cbrcp)$ is closed: it is driven only by the control input $u$ and does not depend on $z$.
Second, the iodine and xenon concentrations $(\I, \Xe)$ are driven by the power $\P$, but with time constants of several hours:
they act as low-pass filters of the fast variables, so that over the $30$-minute prediction horizon of the NMPC,
the influence of the fast transients of $z$ on $x$ remains limited.
Conversely, $z$ depends strongly on $x$, through the reactivity $\rho$, which depends on the boron concentration and on the xenon.
Since $\varepsilon \ll 1$, the fast state quickly converges to a quasi-steady state
$z \approx z^*(x, u, w)$ parametrized by the slow state, and the remaining information carried by $z$ mostly describes fast transients.
Encoding each timescale separately thus reflects the singularly perturbed structure of~\eqref{eq:full_dynamics},
and leaves the coupling between the slow manifold and the fast transients to the controller.

\paragraph{Explicit interaction terms.}
Intermediate architectures, in which explicit interaction terms between the latent blocks are added to the encoders
(e.g., bilinear terms or cross-attention between $\phi_x$, $\phi_z$ and $\phi_w$), are a natural extension of this work.
Such terms could encode the known one-way couplings of the dynamics, e.g., from $x$ to $z$ through the reactivity,
while preserving the attribution of each latent component to a group of physical variables.
We leave this study for future work.

\section{Evaluation Checklist}

\begin{enumerate}
    \item \textbf{Physics feasibility.}
    We assess feasibility by verifying that all trajectories generated using the controller satisfy the operational and safety constraints of the nuclear reactor model. 
    Constraint-violation statistics are reported in Section~\ref{sec:Model}, Table~\ref{tab:quantitative_results}.

    \item \textbf{Out-of-distribution evaluation.}
    The test trajectories are generated independently from the training data and therefore correspond to unseen $48$-hour long load-following scenarios. 
    However, they are sampled from the same underlying distribution as the training trajectories and therefore do not constitute a strict out-of-distribution evaluation. 
    The test set instead assesses generalization to unseen realizations within the training distribution.

    \item \textbf{Failure modes.}
    Failure cases are quantified through feasibility violation statistics reported in Table~\ref{tab:quantitative_results}. 
    While the magnitude of constraint violations could provide additional information, it is of secondary importance in this application.
    The primary requirement is binary feasibility: a load-following trajectory must remain within the admissible operational and safety domain.
    In most cases, the violations remain relatively small.
    A representative failure case illustrating the evolution of the violated safety variable is provided in Appendix~\ref{appendix:example}.

    \item \textbf{System-level effect.}
    The presented controllers are evaluated both as standalone controllers and as warmstarts for the NMPC expert controller. 
    At the system level, their integration into the NMPC pipeline is assessed in terms of computational cost and feasibility. 
    The ML warmstart reduces the NMPC computation time while preserving the feasibility of the resulting control trajectories (Section~\ref{sec:Model}, Table~\ref{tab:quantitative_results}).

    \item \textbf{Data and code.}
    The code used in this work is proprietary and cannot be publicly released due to industrial confidentiality constraints. 
    The underlying reactor simulation model and associated datasets are also subject to the same restrictions. 
    However, algorithmic details, experimental protocols, and aggregate metrics are fully reported in the paper.
    No release of the proprietary code or underlying dataset is planned.
\end{enumerate}

\end{document}